\documentclass[10pt]{article}
\usepackage{times}
\usepackage{etoolbox}

\usepackage[utf8]{inputenc}
\usepackage[T1]{fontenc}
\usepackage{graphicx}
\usepackage{booktabs}
\usepackage{caption}
\usepackage{subcaption}
\usepackage{array}
\usepackage{float}
\usepackage{hyperref}
\usepackage{url}
\usepackage{relsize}
\usepackage{amssymb}
\usepackage{xcolor,pifont}
\usepackage[table,xcdraw]{xcolor}
\usepackage{wrapfig}
\usepackage{amsmath}
\newcommand{\proposed}{\texttt{M\textsuperscript{2}G-LLM }}
\newcommand{\xmark}{\text{\ding{55}}}
\usepackage{jabbrv}
\usepackage[margin=0.75in]{geometry}

\title{M$^2$G-LLM: Enhancing Clinical Prediction via Multimodal Graph Reasoning and LLM Context Injection}

\author{%
  Inyoung Choi$^1$, Sukwon Yun$^2$, Jiayi Xin$^1$, Jie Peng$^2$, Tianlong Chen$^2$, Qi Long$^{1*}$ 
  \\
  \noalign{\vskip 1ex}
  $^1$University of Pennsylvania, Philadelphia, PA, USA \\
  $^2$University of North Carolina at Chapel Hill, Chapel Hill, NC, USA \\
  \noalign{\vskip 1ex}
  \small *Corresponding author: \texttt{qlong@upenn.edu} \\
  \small \texttt{inyoungc@seas.upenn.edu}
}
\begin{document}

\flushbottom
\date{}
\maketitle
% \thispagestyle{empty}

\begin{abstract}
Integrating diverse data modalities—such as clinical notes, laboratory results, and medical imaging—is essential for advancing clinical decision-making. While Large Language Models (LLMs) have shown remarkable performance in processing unstructured clinical text, their limited capacity to incorporate non-text modalities hinders their broader utility in healthcare applications. Here, we introduce \proposed  (Multimodal MedGraph-LLM), a novel framework that enhances LLMs with multimodal integration and alignment via Graph Neural Networks (GNNs). Our approach models temporal relationships between patient visits, propagates information across clinically similar patients, and aligns heterogeneous data sources to construct enriched multimodal context vectors. These vectors are injected into the intermediate layers of the LLM, enabling joint reasoning over textual and non-textual modalities. We evaluate \proposed on the MIMIC-IV and MIMIC-CXR datasets, demonstrating improvements in clinical prediction tasks over strong baseline models. Our results highlight the promise of combining the language understanding of LLMs with the relational reasoning capabilities of GNNs for comprehensive, multimodal healthcare analysis.
\end{abstract}

\section*{Introduction}

Electronic Health Records (EHRs) are a rich source of multimodal data, encompassing unstructured clinical narratives, structured diagnostic codes (e.g., ICD-9/10), laboratory values, and imaging data \cite{hoerbst2010electronic}. The fusion of these diverse modalities offers the potential to enhance clinical modeling and decision support by enabling a more holistic view of patient health \cite{huang2020clinicalbert, kim2022medicalcodes, zhang2019biowordvec, xu2021mufasa, yang2021multimodal, MMbetter}. However, existing approaches often struggle to effectively integrate these heterogeneous data types, particularly in settings that demand longitudinal modeling and personalized reasoning.

Large Language Models (LLMs) have demonstrated exceptional capabilities in processing and reasoning over clinical text. Yet, when deployed in real-world healthcare settings, their utility remains constrained by their limited ability to incorporate structured and non-textual data \cite{nazi2024largemodels, zhou2024surveylargelanguagemodels, kim2022medicalcodes}. Designed primarily for natural language, LLMs are not inherently equipped to integrate tabular, coded, or imaging modalities. Moreover, they often operate in isolation, processing a single patient’s data at a single time point, without access to the broader clinical context that informs medical decision-making.

Efforts to extend LLMs for multimodal healthcare tasks have introduced techniques such as embedding alignment and external memory retrieval \cite{song2023bridgegapmodalitiescomprehensive, he2024surveylargelanguagemodels}. While promising, many recent medical Multimodal LLM (MLLM) architectures, such as radiology Vision-Language Models (VLMs), focus primarily on aligning medical imaging with text via projection layers. These methods—including LLaVA-Med, Med-Flamingo, and RadFM—frequently rely on 'pseudo-token' injection, where non-text modalities are projected into the LLM's input stream as linguistic tokens \cite{li2023llavamedtraininglargelanguageandvision, moor2023medflamingomultimodalmedicalfewshot, wu2023generalistfoundationmodelradiology}. While effective for single-encounter tasks, these episodic approaches suffer from sample isolation and temporal blindness, failing to capture the longitudinal progression of patient trajectories or the similarity structure across broader clinical populations. Furthermore, concatenation-based fusion strategies scale poorly with long patient histories, often exhausting the LLM's context window \cite{Huang_2023, mumtaz2024llmshealthcarecurrentapplications, Yildirim_2024}.

To address these limitations, we introduce \textbf{\proposed} (Figure \ref{fig:main_figure}), a unified framework that combines the representational power of LLMs with the relational structure of Graph Neural Networks (GNNs) and the flexibility of contrastive alignment techniques. ~\proposed constructs patient-level graphs to encode temporal and similarity relationships, learns modality-specific embeddings aligned in a shared space, and integrates this information into the LLM via residual-stream conditioning \cite{li2024implicitincontextlearning}.  The novelty of our approach lies not in the injection primitive itself—residual-stream conditioning is adopted from prior work \cite{li2024implicitincontextlearning}-but in the integration pattern it enables: to our knowledge, \proposed is the first framework to derive a cross-patient, longitudinally-informed context vector from a population graph and use it to condition an LLM at depth. This distinguishes it from two dominant paradigms. Unlike pseudo-token approaches (e.g., HeLM, LLaVA-Med), which project non-text modalities into the input stream and consume context-window capacity, \proposed conditions the residual stream and leaves the prompt untouched—an advantage as patient histories grow long. Unlike prefix- and soft-prompt tuning, which learn a fixed set of task parameters shared across all inputs, our context vector carries no free parameters of its own: it is computed per patient as a function of that patient's record and of clinically similar patients, making it an input-dependent representation rather than a learned constant. This design enables ~\proposed to reason over multimodal patient histories and produce more context-aware and clinically relevant predictions—without altering the underlying LLM architecture \cite{Zhang_2022_m3care, bommasani2022opportunitiesrisksfoundationmodels, hager2024evaluation, Ageno2023}.

\section*{Results}

\begin{table*}[!t]
\centering
\caption{Performance of all models on one-year mortality prediction with 95\% confidence intervals. \proposed consistently outperforms baselines on this task.}
\label{tab:mortality_results}
% --- Unified Performance Table with 95% Confidence Intervals ---
\renewcommand{\arraystretch}{1.3}
\resizebox{\textwidth}{!}{
\begin{tabular}{lcccccccccc}
\toprule
& \textbf{LLM-based} & \textbf{Lab+Vital} & \textbf{Code} & \textbf{Image} & \textbf{Text} & \textbf{Acc} & \textbf{F1} & \textbf{AUC} & \textbf{AUPRC} & \textbf{Balanced Accuracy}\\
\toprule
HAIM & \small \xmark & \large \checkmark & \large \checkmark & \large \checkmark & \large \checkmark & \small 71.24 \scriptsize{(70.82--71.66)} & \small 58.38 \scriptsize{(58.11--58.65)} & \small 68.48 \scriptsize{(68.16--68.80)} & \small 32.34 \scriptsize{(31.47--33.21)} & \small 59.90 \scriptsize{(59.12--60.68)} \\
M3Care & \small \xmark & \large \checkmark & \large \checkmark & \large \checkmark & \large \checkmark & \small 77.60 \scriptsize{(76.75--78.45)} & \small 55.30 \scriptsize{(54.46--56.14)} & \small 67.31 \scriptsize{(66.68--67.94)} & \small 31.42 \scriptsize{(30.69--32.15)} & \small 57.08 \scriptsize{(56.28--57.88)} \\
MUSE & \small \xmark & \large \checkmark & \large \checkmark & \small \xmark & \large \checkmark & \small 77.21 \scriptsize{(76.24--78.18)} & \small 51.89 \scriptsize{(50.64--53.14)} & \small 65.27 \scriptsize{(64.45--66.09)} & \small 31.21 \scriptsize{(29.74--32.68)} & \small 53.98 \scriptsize{(52.73--55.23)} \\
mmFormer & \small \xmark & \large \checkmark & \large \checkmark & \large \checkmark & \large \checkmark & \small 76.28 \scriptsize{(75.48--77.08)} & \small 62.34 \scriptsize{(61.39--63.29)} & \small 69.27 \scriptsize{(67.84--70.70)} & \small 32.54 \scriptsize{(30.83--34.25)} & \small 62.56 \scriptsize{(61.76--63.36)} \\
HeLM & \large \checkmark & \large \checkmark & \small \xmark & \small \xmark & \large \checkmark & \small 74.89 \scriptsize{(73.69--76.09)} & \small 51.12 \scriptsize{(50.20--52.04)} & \small 55.42 \scriptsize{(54.58--56.26)} & \small 29.81 \scriptsize{(28.83--30.79)} & \small 55.32 \scriptsize{(54.45--56.19)} \\
LLMM & \large \checkmark & \large \checkmark & \small \xmark & \small \xmark & \large \checkmark & \small 75.35 \scriptsize{(75.02--75.68)} & \small 48.73 \scriptsize{(48.14--49.32)} & \small 54.57 \scriptsize{(54.28--54.86)} & \small 24.49 \scriptsize{(23.46--25.52)} & \small 50.10 \scriptsize{(49.78--50.42)} \\
\mbox{Rezk et al., 2024} & \large \checkmark & \large \checkmark & \large \checkmark & \small \xmark & \large \checkmark & \small 74.99 \scriptsize{(74.31--75.67)} & \small \textbf{65.56} \scriptsize{(64.38--66.74)} & \small 62.96 \scriptsize{(61.66--64.26)} & \small 31.13 \scriptsize{(30.52--31.74)} & \small 60.43 \scriptsize{(59.57--61.29)} \\
Llama 3-8b & \large \checkmark & \small \xmark & \small \xmark & \small \xmark & \large \checkmark & \small 69.80 \scriptsize{(69.54--70.06)} & \small 49.85 \scriptsize{(49.49--50.21)} & \small 55.81 \scriptsize{(54.83--56.79)} & \small 21.78 \scriptsize{(21.29--22.27)} & \small 50.87 \scriptsize{(49.82--51.92)} \\
LLaVA & \large \checkmark & \small \xmark & \small \xmark & \small \checkmark & \large \checkmark & \small 65.75 \scriptsize{(64.81--66.69)} & \small 55.17 \scriptsize{(54.68--55.66)} & \small 56.65 \scriptsize{(56.07--57.23)} & \small 27.47 \scriptsize{(26.51--28.43)} & \small 56.19 \scriptsize{(55.34--57.04)} \\
\textbf{~\proposed } & \large \checkmark & \large \checkmark & \large \checkmark & \large \checkmark & \large \checkmark & \textbf{78.98} \scriptsize{(78.17--79.79)} & 64.55 \scriptsize{(63.53--65.57)} & \textbf{70.22} \scriptsize{(68.59--71.85)} & \textbf{34.15} \scriptsize{(32.51--35.79)} & \textbf{63.40} \scriptsize{(61.74--65.06)} \\
\bottomrule
\end{tabular}
}
\end{table*}
\vspace{2pt}
\begin{table*}[!t]
\caption{Performance of all models on 30-day readmission prediction tasks with 95\% confidence intervals.}
\label{tab:readmission_results}
\renewcommand{\arraystretch}{1.3}
\resizebox{\textwidth}{!}{
\begin{tabular}{lcccccccccc}
\toprule
& \textbf{LLM-based} & \textbf{Lab+Vital} & \textbf{Code} & \textbf{Image} & \textbf{Text} & \textbf{Acc} & \textbf{F1} & \textbf{AUC} & \textbf{AUPRC} & \textbf{Balanced Accuracy}\\
\toprule
HAIM & \small \xmark & \large \checkmark & \large \checkmark & \large \checkmark & \large \checkmark & \small 66.74 \scriptsize{(66.14--67.34)} & \small 48.54 \scriptsize{(47.50--49.58)} & \small 48.08 \scriptsize{(47.40--48.76)} & \small 13.38 \scriptsize{(13.05--13.71)} & \small 49.26 \scriptsize{(48.27--50.25)} \\
M3Care & \small \xmark & \large \checkmark & \large \checkmark & \large \checkmark & \large \checkmark & \small 80.57 \scriptsize{(79.02--82.12)} & \small 49.57 \scriptsize{(49.15--49.99)} & \small 51.14 \scriptsize{(50.27--52.01)} & \small 15.88 \scriptsize{(15.43--16.33)} & \small 49.83 \scriptsize{(48.58--51.08)} \\
MUSE & \small \xmark & \large \checkmark & \large \checkmark & \small \xmark & \large \checkmark & \small 76.18 \scriptsize{(75.71--76.65)} & \small 49.24 \scriptsize{(48.82--49.66)} & \small 46.22 \scriptsize{(45.96--46.48)} & \small 12.45 \scriptsize{(11.99--12.91)} & \small 48.23 \scriptsize{(47.64--48.82)} \\
mmFormer & \small \xmark & \large \checkmark & \large \checkmark & \large \checkmark & \large \checkmark & \small 85.04 \scriptsize{(84.40--85.68)} & \small 46.82 \scriptsize{(46.23--47.41)} & \small 49.52 \scriptsize{(48.19--50.85)} & \small 15.65 \scriptsize{(15.08--16.22)} & \small 48.66 \scriptsize{(48.10--49.22)} \\
HeLM & \large \checkmark & \large \checkmark & \small \xmark & \small \xmark & \large \checkmark & \small 71.89 \scriptsize{(70.89--72.89)} & \small 50.44 \scriptsize{(49.41--51.47)} & \small 51.94 \scriptsize{(51.28--52.60)} & \small 17.12 \scriptsize{(16.48--17.76)} & \small 50.49 \scriptsize{(49.56--51.42)} \\
LLMM & \large \checkmark & \large \checkmark & \small \xmark & \small \xmark & \large \checkmark & \small 79.27 \scriptsize{(78.40--80.14)} & \small 47.69 \scriptsize{(47.46--47.92)} & \small 50.11 \scriptsize{(49.81--50.41)} & \small 15.12 \scriptsize{(14.93--15.31)} & \small 50.05 \scriptsize{(49.38--50.72)} \\
\mbox{Rezk et al., 2024} & \large \checkmark & \large \checkmark & \large \checkmark & \small \xmark & \large \checkmark & \small 72.08 \scriptsize{(71.70--72.46)} & \small 50.62 \scriptsize{(49.74--51.50)} & \small 49.53 \scriptsize{(48.61--50.45)} & \small 14.13 \scriptsize{(12.97--15.29)} & \small 49.36 \scriptsize{(48.48--50.24)} \\
Llama 3-8b & \large \checkmark & \small \xmark & \small \xmark & \small \xmark & \large \checkmark & \small 74.78 \scriptsize{(74.45--75.11)} & \small 48.66 \scriptsize{(47.80--49.52)} & \small 53.65 \scriptsize{(52.74--54.56)} & \small 17.83 \scriptsize{(17.14--18.52)} & \small 50.71 \scriptsize{(49.76--51.66)} \\
LLaVA & \large \checkmark & \small \xmark & \small \xmark & \small \checkmark & \large \checkmark & \small 66.82 \scriptsize{(66.22--67.42)} & \small 49.82 \scriptsize{(49.55--50.09)} & \small 50.42 \scriptsize{(50.20--50.64)} & \small 16.83 \scriptsize{(16.48--17.18)} & \small 49.56 \scriptsize{(48.97--50.15)} \\
\textbf{~\proposed } & \large \checkmark & \large \checkmark & \large \checkmark & \large \checkmark & \large \checkmark & \textbf{85.75} \scriptsize{(85.45--86.05)} & \textbf{52.35} \scriptsize{(51.47--53.23)} & \textbf{54.80} \scriptsize{(53.98--55.62)} & \textbf{20.15} \scriptsize{(18.98--21.32)} & \textbf{51.20} \scriptsize{(50.12--52.28)} \\
\bottomrule
\end{tabular}
}
\end{table*}

We evaluate ~\proposed on two clinical prediction tasks using the MIMIC-IV and MIMIC-CXR datasets: (1) one-year mortality prediction and (2) 30-day readmission. Our results demonstrate that ~\proposed outperforms both LLM-based and non-LLM multimodal baselines across standard metrics, while also offering natural-language explanations and degrading more gracefully than baselines when modalities are missing at inference.

\subsection*{~\proposed Achieves State-of-the-Art Performance in Clinical Prediction} Table ~\ref{tab:mortality_results} and Table ~\ref{tab:readmission_results} compare ~\proposed to a comprehensive set of baselines on the two primary tasks. ~\proposed achieves the highest overall performance on both tasks. For one-year mortality prediction, ~\proposed yields 78.98\% accuracy and 64.55 F1 score, outperforming both traditional multimodal models and LLM-based approaches. Similarly, on the readmission task, ~\proposed attains 85.75\% accuracy and 52.35 F1, again surpassing competitive baselines. These gains are consistent across 40 independent runs and reflected in robust AUC metrics, highlighting the generalizability of ~\proposed across different clinical endpoints.

LLM-based baselines that incorporate limited modality fusion (e.g., LLMM, HeLM) demonstrate moderate performance improvements over text-only models, but underperform relative to ~\proposed. Notably, approaches such as Rezk et al., which convert structured inputs into text, exhibit improved F1 scores but suffer from lower accuracy, suggesting an imbalance in capturing minority class distributions. In contrast, ~\proposed leverages GNNs to integrate multimodal signals structurally and temporally, leading to more balanced and clinically actionable predictions.

Additional analyses of model reliability and fairness, including Brier scores, Expected Calibration Error, and performance across demographic subgroups, are provided in the Supplementary Information (Supplementary Tables 1-7) ; M²G-LLM demonstrated superior calibration and stability compared to all baselines.

\begin{figure*}[ht]
\centering
\includegraphics[width=\linewidth]{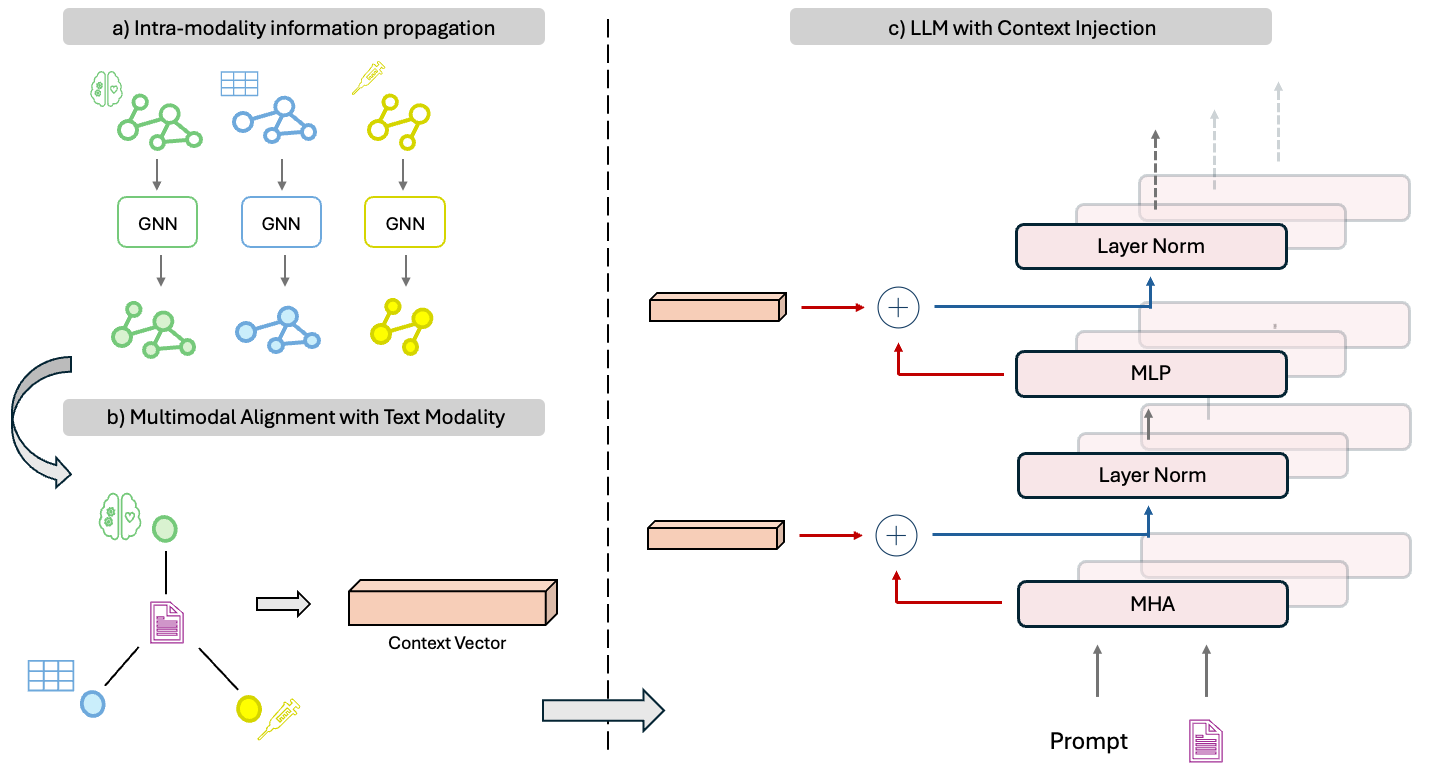}
\caption{}
\label{fig:main_figure}
\end{figure*}

\subsection*{All Modalities Matter} 
\vspace{-0.7em}
To better understand the contribution of each modality, we perform a comprehensive ablation study (Table~\ref{tab:ablation_modalities}). The removal of any individual modality results in a consistent degradation in predictive performance, underscoring the importance of each data source. Among all modalities, structured data—specifically laboratory measurements and diagnosis codes—proves to be most impactful. When lab values are excluded, the F1 score drops to 58.34, a substantial decline that suggests the role of laboratory results in clinical risk prediction. 

Interestingly, models utilizing only clinical text and imaging data perform worst (F1 score of 53.12), highlighting that visual and narrative information alone cannot fully capture the nuances of patient physiology or disease progression. This observation aligns with clinical intuition: while radiology reports and notes offer rich contextual information, they are often subjective or delayed representations of underlying pathophysiology—making structured signals like lab abnormalities indispensable for timely and accurate prediction.

Conversely, the integration of all four modalities—clinical notes, lab results, diagnosis codes, and imaging—yields the best predictive performance (F1 score of 64.55), validating the premise that ~\proposed benefits from synergistic multimodal reasoning rather than dominance by a single modality. This synergy is made possible through our alignment and propagation framework, which ensures that complementary modalities reinforce each other rather than compete for representational space. The results are reported across 40 runs and ablation results for the 30-day readmission task are included in the Supplementary Materials (Supplementary Tables 8-9).

\subsection*{Graph Propagation and Multimodal Alignment Drive Performance}
\vspace{-0.7em}
We assess the contribution of ~\proposed's core architectural components through an ablation study in which key modules are frozen during training, and we report the resulting performance changes. Specifically, we isolate the effects of (1) graph-based propagation and (2) multimodal alignment using contrastive learning. Results are shown in Table \ref{tab:frozen}.

Freezing the graph propagation module, which prevents it from learning task-specific propagation while retaining its randomly initialized transformation, results in a performance decrease from 64.55 F1 to 59.72 F1. The accuracy similarly drops from 78.98 to 76.41. These results indicate that the inclusion of GNN-based information propagation improves predictive performance on clinical tasks. Without propagation, embeddings from different visits and patients remain independent, which appears to limit the model’s ability to integrate broader patient context.

Freezing the modality alignment module leads to a more pronounced decline, with F1 decreasing to 47.21. This component is responsible for mapping non-text modalities (labs, codes, and imaging) into a shared latent space centered around the clinical text. The drop in performance when this module is frozen suggests that embedding alignment plays a substantial role in facilitating effective multimodal fusion.

We also compare these results with the performance of the full ~\proposed model, which includes both graph propagation and alignment. This version achieves the highest performance, with 78.98 accuracy and 64.55 F1, indicating that both components contribute complementary gains.

Together, the results from these ablations demonstrate that both graph propagation and multimodal alignment provide measurable benefits. Disabling either one leads to a substantial decrease in model performance across evaluation metrics. Notably, the alignment component appears to have the larger individual impact. These empirical findings support the inclusion of both modules as necessary components of the \proposed architecture for robust multimodal healthcare prediction.

Because complete multimodal data is rarely available at decision time, we evaluate performance when modalities are missing for individual patients at inference rather than withheld during training. Dropping each structured modality independently per patient with probability p, M²G-LLM declines from 0.702 to 0.669 AUC as p rises from 0 to 0.8, compared with a decline from 0.693 to 0.587 for mmFormer. Full results are given in Supplementary Section S9. This graceful degradation is consistent with the reconstruction mechanism described in Methods: because missingness is applied at the GNN input, message passing recovers a non-zero representation for an absent modality from a node's clinically similar neighbors.

Because freezing preserves the module in the architecture, we additionally evaluate a graph-free control in which the GNN modules are removed entirely and the initial modality embeddings are mapped directly into the shared space, with fusion, alignment, and injection otherwise identical and trained under the same procedure. This isolates cross-patient graph construction as the only difference between the two models. Removing the graph reduces one-year mortality AUC from 0.702 to 0.685 and 30-day readmission AUC from 0.548 to 0.527 (Supplementary Table 12), indicating that performance is not attributable to multimodal fusion and residual-stream injection alone.

\subsection*{Alignment Yields Unified Embedding Space
} 
\begin{figure}[ht]
    \centering
    \includegraphics[width=0.95\textwidth]{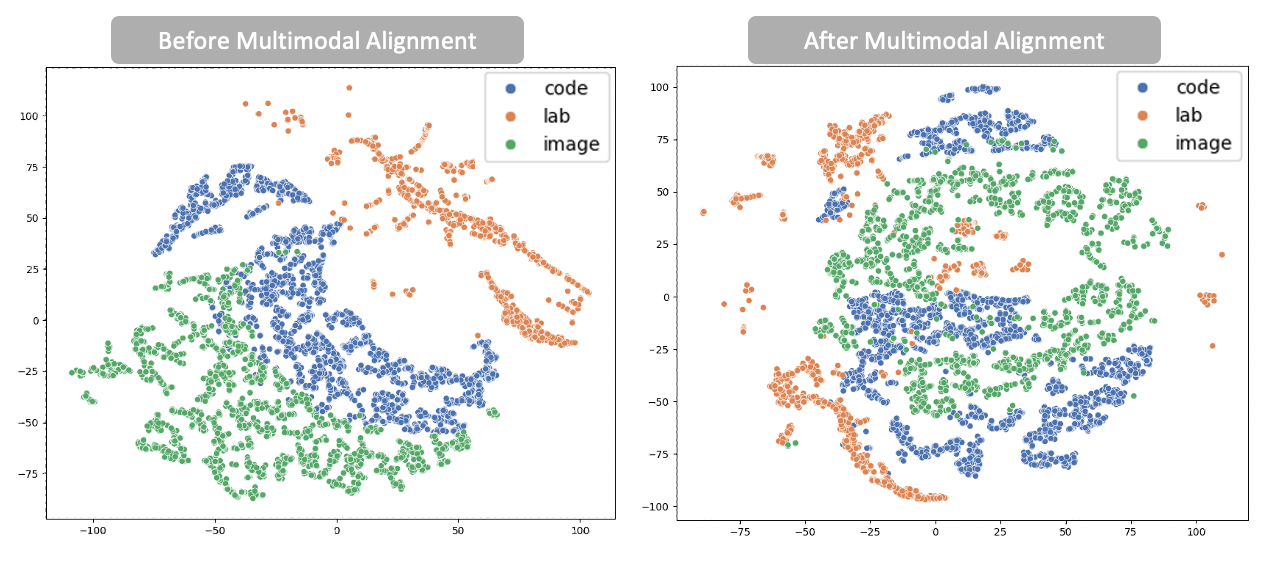}
    \caption{}
    \label{fig:tsne_alignment}
\end{figure}

\vspace{-0.7em}
To evaluate the effect of the multimodal alignment module, we visualize modality-specific embeddings before and after alignment using t-SNE projection. Figure ~\ref{fig:tsne_alignment} presents two-dimensional t-SNE plots of the embeddings from clinical text, lab values, diagnosis codes, and imaging data.

Prior to alignment, embeddings from each modality form distinct, non-overlapping clusters. Text, lab, code, and image embeddings are clearly separated, with minimal overlap in the projection space. This indicates that without alignment, each modality resides in its own disjoint feature space, making direct interaction between them within the LLM challenging.

After applying contrastive alignment, the updated embeddings display a noticeably different spatial distribution. The modality-specific clusters become more interspersed, with embeddings from different modalities appearing closer together in the projected space. This change reflects a measurable shift toward a more cohesive and unified embedding space. This empirical observation demonstrates that the alignment procedure reduces modality-specific variance and increases cross-modal proximity. By comparing the t-SNE plots before and after alignment, we confirm that the contrastive training objective effectively reorganizes the representation space to support subsequent integration within the LLM.

\subsection*{\proposed Produces Interpretable Explanations in Natural Language} 
\vspace{-0.7em}
In addition to quantitative improvements in prediction accuracy, ~\proposed is designed to generate free-text explanations alongside its output. To evaluate this capability, we collect model-generated explanations for both positive and negative one-year mortality predictions and analyze representative outputs. Table ~\ref{tab:qualitative_explanations} displays sample explanations corresponding to both high-risk and low-risk prediction cases.

For patients predicted to be at high risk of mortality, ~\proposed consistently produces explanations that reference relevant clinical findings. For example, in one case, the model highlights a history of congestive heart failure and elevated BUN/creatinine ratios observed across multiple visits. In another, the explanation notes frequent admissions, progressive respiratory failure, and abnormal radiology findings (e.g., bilateral effusion). These outputs reflect the model’s ability to surface temporally distributed clinical features, such as comorbidities and abnormal measurements, and present them in coherent narrative form.

In low-risk predictions, the generated explanations reference indicators of clinical stability. Examples include phrases such as “vital signs remained within normal range,” “blood glucose levels well-managed,” or “no major procedures or diagnoses recorded in recent visits.” These explanations often span multiple modalities, referencing lab values, diagnosis code trends, and unremarkable imaging results, indicating that the model is drawing on the multimodal context vector injected during inference.

\begin{table*}[t]
\centering
\begin{minipage}[t]{0.48\textwidth}
\centering
\resizebox{\textwidth}{!}{ 
\begin{tabular}{lllllll}
\toprule
\textbf{Lab+Vital} & \textbf{Code} & \textbf{Image} & \textbf{Text} & \textbf{Acc} & \textbf{F1} & \textbf{AUC} \\
\midrule
{\xmark}    & {\checkmark}    & {\xmark}    & {\checkmark}    & 75.86 \scriptsize{(75.75--75.97)} & 56.45 \scriptsize{(56.09--56.81)} & 64.57 \scriptsize{(64.38--64.76)} \\
{\checkmark}    & {\xmark}    & {\xmark}    & {\checkmark}    & 76.39 \scriptsize{(76.18--76.60)} & 57.80 \scriptsize{(57.61--57.99)} & 64.41 \scriptsize{(64.28--64.54)} \\
{\xmark}    & {\xmark}    & {\checkmark}    & {\checkmark}    & 74.46 \scriptsize{(74.20--74.72)} & 53.12 \scriptsize{(52.92--53.32)} & 61.58 \scriptsize{(61.41--61.75)} \\
{\xmark}    & {\checkmark}    & {\checkmark}    & {\checkmark}    & 77.91 \scriptsize{(77.72--78.10)} & 58.34 \scriptsize{(58.18--58.50)} & 66.92 \scriptsize{(66.79--67.05)} \\
{\checkmark}    & {\xmark}    & {\checkmark}    & {\checkmark}    & 76.48 \scriptsize{(76.37--76.59)} & 59.49 \scriptsize{(59.35--59.63)} & 66.55 \scriptsize{(66.45--66.65)} \\
{\checkmark}    & {\checkmark}    & {\xmark}    & {\checkmark}    & 77.34 \scriptsize{(77.21--77.47)} & 60.42 \scriptsize{(60.10--60.74)} & 68.72 \scriptsize{(68.57--68.87)} \\
\midrule
{\checkmark}    & {\checkmark}    & {\checkmark}    & {\checkmark}    & \textbf{78.98} \scriptsize{(78.17--79.79)} & \textbf{64.55} \scriptsize{(63.53--65.57)} & \textbf{70.22} \scriptsize{(68.59--71.85)} \\
\bottomrule
\end{tabular} 
}
\caption{Ablation study of data modalities for one-year mortality prediction.}
\label{tab:ablation_modalities} 
\end{minipage}
\hfill
% --- SECOND TABLE ---
\begin{minipage}[t]{0.48\textwidth}
\centering
\resizebox{\textwidth}{!}{ 
\begin{tabular}{lccc}
\toprule
\textbf{Frozen Component} & \textbf{Accuracy} & \textbf{F1} & \textbf{AUC} \\
\midrule
Graph Propagation          & 76.41 \scriptsize{(76.29--76.53)} & 59.72 \scriptsize{(59.53--59.91)} & 61.84 \scriptsize{(61.01--62.67)} \\
Multimodal Alignment       & 74.89 \scriptsize{(74.59--75.19)} & 47.21 \scriptsize{(47.10--47.32)} & 52.89 \scriptsize{(52.79--52.99)} \\
\midrule
Full Model (M$^2$G-LLM)   & \textbf{78.98} \scriptsize{(78.17--79.79)} & \textbf{64.55} \scriptsize{(63.53--65.57)} & \textbf{70.22} \scriptsize{(68.59--71.85)} \\
\bottomrule
\end{tabular}
}

\caption{Architectural ablation results for the one-year mortality task obtained by freezing key framework components during training.}
\label{tab:frozen}
\end{minipage}

\end{table*}

\section*{Discussion}
This work introduces ~\proposed, a unified framework that enhances large language models  with structured multimodal reasoning capabilities for clinical prediction. By integrating unstructured text, structured codes and labs, and imaging data through graph-based propagation and contrastive alignment, ~\proposed bridges the gap between the language reasoning capabilities of LLMs and the multimodal, longitudinal structure of EHRs. Our results demonstrate consistent improvements over both LLM-centric and traditional multimodal baselines on critical clinical tasks, including one-year mortality and 30-day readmission prediction. Beyond performance gains, ~\proposed offers compelling advances in transparency, flexibility, and robustness—key pillars for clinical adoption of AI systems.

% \subsection*{Beyond Text: Toward Truly Multimodal Clinical LLMs}
While LLMs have revolutionized clinical natural language understanding, they remain inherently limited by their reliance on text-based inputs. Healthcare data, however, is inherently multimodal—ranging from high-dimensional imaging to sparse structured codes and dense numeric lab values. ~\proposed is explicitly designed to overcome this limitation by embedding non-textual data into a shared latent space and injecting this information directly into the LLM's internal representations.

This approach marks a departure from prior work that relies on serializing structured data into text or concatenating modality-specific embeddings. Instead, ~\proposed incorporates a patient-specific multimodal context vector into the residual stream of the LLM, allowing it to preserve its pretrained language modeling capabilities while being conditioned on a rich, cross-modal representation of patient state. This also supports prediction for patients with heterogeneous data availability: because missingness is handled at the graph input, a modality absent for a given patient–visit is reconstructed from that node's neighbourhood rather than propagated forward as an empty representation (Supplementary Section S9).

By modeling both temporal visit trajectories and inter-patient similarities, the graph-based module enriches the LLM’s understanding of longitudinal and population-level patterns—dimensions of reasoning that are central to clinical decision-making but inaccessible to text-only models. ~\proposed thus represents a step toward truly context-aware, temporally grounded, and population-informed LLMs for healthcare.

% \subsection*{Clinical Relevance and Interpretability}
One of the most compelling features of ~\proposed is its capacity to generate interpretable natural language explanations for its predictions. In clinical domains, model transparency is essential for adoption. By aligning heterogeneous modalities and conditioning LLM outputs on a shared, patient-level representation, ~\proposed provides explanations that are both medically coherent and patient-specific. These explanations not only increase trust but may also facilitate hypothesis generation or decision support in real-world applications. Moreover, our t-SNE visualizations illustrate how the modality alignment module projects disparate modalities into a unified space, enabling consistent downstream reasoning. The ablation studies further emphasize that each modality adds distinct, complementary value, particularly structured data such as labs and diagnosis codes, which are underutilized in many LLM-based approaches.

% \subsection*{Limitations and Future Directions}
Despite the promising results, \proposed\ has limitations that inform future directions. First, we quantify the framework's computational cost (Supplementary Section S11). On a single NVIDIA B200, training requires 492 s per epoch over approximately five epochs, and inference costs 154 ms per patient, with peak memory of 36.1 GB during a 4096-token forward pass; the Llama-3-8B backbone occupies 16.3 GB in bfloat16, and only 53.11 M parameters (0.66\% of 8.03 B) are trained. Similarity-edge construction is O(N²d) but is incurred once offline, and no similarity computation occurs at training or inference time. The binding constraint on scale is therefore memory rather than latency: the dense N × N similarity matrix requires roughly 866 MB in float32 at the present cohort size and grows quadratically.

Second, while \proposed demonstrates strong performance on retrospective prediction tasks using MIMIC-IV and MIMIC-CXR, we emphasize that this work is primarily designed to assess the methodological contribution of our proposed integration architecture. Our experiments serve as a controlled benchmark to isolate the impact of graph-based relational context when fused with frozen LLMs. Consequently, establishing the downstream generalizability of a specific prediction model derived from this framework—through external validation on independent hospital cohorts or prospective clinical trials—is beyond the immediate scope of this methodological study. Future work will focus on applying this \proposed blueprint to diverse institutional data to evaluate its performance under different clinical coding practices and data sparsity patterns, which remains a critical next step for clinical translation.

Third, while we show that \proposed\ generates faithful explanations in natural language, systematic evaluation of explanation quality—particularly alignment with clinical judgment—remains to be performed. Incorporating clinician-in-the-loop validation and developing quantitative metrics for explanation fidelity will be crucial for building trust in real-world use.

Two aspects of our ablation design merit acknowledgment. We assess the graph and alignment modules by freezing rather than removing them, which preserves a valid end-to-end architecture but does not fully separate architectural contribution from training dynamics; the reported decrements are therefore upper bounds, and we complement them with a graph-free control (Supplementary Table 12) that removes the graph entirely. We likewise treat temporal and similarity edges uniformly and remove them together, so their individual contributions are not isolated — separating them would confound the result with connectivity changes for single-visit patients — and we defer a graded edge-type ablation to future work.

Finally, while \proposed\ handles missing modalities implicitly via its graph-based design, explicit learning under partial modality supervision, or active selection of informative modalities per patient, could further increase efficiency and robustness in multimodal clinical modeling.
% \subsection*{Clinical Relevance and Broader Implications}
\proposed is designed to address practical challenges that limit the clinical translation of multimodal prediction models, including heterogeneous data availability, temporal fragmentation of patient records, and the difficulty of integrating structured EHR data with language-based clinical reasoning. By operating on routinely collected modalities—clinical notes, laboratory results, diagnosis codes, and imaging—and by accommodating missing or partially observed modalities, \proposed aligns with real-world hospital settings where complete multimodal data is rarely available at decision time.

From a clinical perspective, the proposed framework is well suited for risk stratification and decision support tasks that depend on longitudinal patient context, such as identifying patients at elevated risk of mortality or readmission at discharge. The use of a LLM augmented with structured multimodal context enables the model to produce both calibrated risk estimates and natural-language explanations, which may support clinician interpretation, auditability, and trust. Importantly, because the LLM backbone is not fine-tuned, the framework can be adapted across institutions without retraining large language models, lowering barriers to deployment and maintenance.

Beyond the specific tasks studied here, \proposed provides a general blueprint for grounding LLM-based clinical systems in structured patient data without resorting to brittle text serialization or prompt engineering alone. As LLMs continue to be explored for medical decision support, triage, and summarization, frameworks such as \proposed offer a principled pathway toward clinically meaningful, interpretable, and deployable multimodal AI systems.

\section*{Methods}
We present \proposed, a unified framework that integrates structured and unstructured clinical data into a Large Language Model (LLM) via graph-based representation, modality alignment, and residual-stream conditioning as shown in Figure \ref{fig:main_figure}. The method comprises three primary components:

\begin{itemize}
    \item \textbf{Graph-based information propagation} using Graph Neural Networks (GNNs) to model intra- and inter-patient relationships;
    
    \item \textbf{Multimodal embedding alignment} through contrastive learning to unify heterogeneous data types into a shared representation space;
    
    \item \textbf{Contextual integration into the LLM} using residual-stream conditioning, allowing external multimodal signals to modulate LLM predictions.
\end{itemize}

We describe each component in detail below.

\begin{table}[ht]
\centering
\caption{\textbf{Qualitative assessment of M$^2$G-LLM mortality predictions.} Representative model-generated explanations for positive and negative risk predictions.}
\label{tab:qualitative_explanations}
\begin{tabular}{|>{\centering\arraybackslash}m{0.25\textwidth}|m{0.65\textwidth}|}
\hline
\textbf{Output} & \centering\arraybackslash \textbf{Explanations} \\ \hline
Positive Mortality Prediction & The patient has a history of coronary artery disease (CAD), multiple myocardial infarctions (MIs), including a recent NSTEMI, and congestive heart failure (CHF) with a severely reduced ejection fraction (EF) of 15\% as noted in his most recent transthoracic echocardiogram (TTE). He is currently on a heparin bridge to Coumadin with a target INR of 2.5, as well as several cardiac medications, including a statin, an ACE-inhibitor, a beta-blocker, and a diuretic. Although his intake and output (I\&O) balance is stable, his overall clinical status, including fluid and sodium restrictions, reflects the severity of his condition. \\ \hline
Negative Mortality Prediction & This patient is at a relatively low risk of mortality. She is a 45-year-old woman with a history of breast cancer, currently undergoing treatment, as well as well-managed hypertension and GERD. Although she is a candidate for a hysterectomy and oophorectomy due to a diagnosis of ovarian cancer, her overall health is stable, and there are no immediate signs of life-threatening complications. Her vital signs are stable, and her comorbidities, such as hypertension and GERD, are controlled with medication. \\ \hline
\end{tabular}
\end{table}

\textbf{Graph-Based Information Propagation.}
To effectively model intra- and inter-patient relationships, we construct a separate graph for each data modality, including codified data (e.g., diagnosis and procedure codes), laboratory results, and medical imaging. Each node in these graphs represents a patient–visit instance, initialized using modality-specific embeddings derived from pretrained encoders. We use BioClinicalBERT~\cite{alsentzer2019clinicalbert} to generate embeddings for clinical notes and codified data, and ResNet50~\cite{he2015deepresiduallearningimage} pretrained on ImageNet for imaging data. Graph construction follows an inductive protocol. Because the cohort is split at the subject level and graphs are constructed separately over disjoint subject sets, no test-subject node is present at training time and no train–test edge exists by construction; no test-set connection or downstream path can therefore influence the learned representations. At inference, test-subject visits are instantiated as nodes of a separate test-time graph and connected under the identical criterion, so that similarity edges are formed among held-out visits rather than against a stored training gallery. The protocol is thus inductive with respect to training and transductive within the inference cohort. This does not introduce label leakage: edge construction depends only on cosine similarity between pretrained modality embeddings and on temporal adjacency between consecutive visits of the same patient, and outcome labels are never used at any stage. We verify this empirically in Supplementary Section S7, where neither graph exhibits outcome homophily beyond chance.
% Graphs used during training are constructed over training subjects only; no test-subject nodes are present and no train–test edges exist at training time, so no test-set connection influences the learned representations. At inference, test-subject visits are added as new nodes and connected to the fixed training graph.

% \paragraph{Graph construction.}
Graphs are constructed independently for each modality and are not fully connected at any stage. We first add temporal edges connecting consecutive visits from the same patient, capturing longitudinal health trajectories. We then add similarity edges connecting visits across different patients using a hybrid \emph{top-$k$ + cosine threshold} strategy. Specifically, we compute pairwise cosine similarity between visit-level embeddings and, for each node, identify the top-$k$ most similar nodes. From these candidates, a similarity edge is added only if the cosine similarity exceeds a modality-specific threshold. Self-loops and edges connecting visits from the same patient are excluded. This procedure yields sparse graphs (mean similarity degree 31.7, 28.5, and 3.4 for the code, image, and lab graphs respectively; Supplementary Table 11) while remaining computationally tractable. We evaluate the clinical content of the resulting neighborhoods directly (Supplementary Section S7). Similarity edges in the codified graph join visits sharing a primary ICD chapter at 2.8× the rate expected under a degree-matched random-edge null, replicating on the held-out split (2.7×), and the laboratory graph is moderately homophilous (1.6–1.8×). No graph is outcome-homophilous: mortality and readmission lifts fall between 0.89 and 1.02 across all modalities and splits, confirming that a construction which never observes labels provides no shortcut to the prediction target.

% \paragraph{Graph neural network architecture.}
Each modality-specific graph is processed using a three-layer Graph Convolutional Network (GCN) implemented with \texttt{GCNConv} from PyTorch Geometric. Given a graph $G = (V, E)$, node features are updated as:
\begin{equation}
    h_v^{(l+1)} = \sigma\left(\sum_{u \in \mathcal{N}(v)} \frac{1}{c_{vu}} W^{(l)} h_u^{(l)}\right),
\end{equation}
where $h_v^{(l)}$ denotes the embedding of node $v$ at layer $l$, $\mathcal{N}(v)$ is its neighborhood, $c_{vu}$ is a normalization factor, and $W^{(l)}$ is a learnable weight matrix.

We perform three rounds of message passing, such that after three GCN layers each node aggregates information from its 3-hop neighborhood. We selected this depth empirically. Supplementary Section S10 compares one-, two-, and three-layer propagation across GCN, GraphSAGE, and GAT: one-year mortality AUC improves monotonically with depth for all three operators (GCN 0.664 → 0.686 → 0.702), while differences between operators at a given depth are small relative to their dispersion (GAT is marginally ahead at each depth, 0.705 at depth three). We retain GCN for parsimony and lower computational cost. We note that returns had not saturated at three layers; depth was capped by the cost of retraining the full framework, and deeper propagation may yield further gains. The input, hidden, and output dimensions are kept identical within each modality-specific GNN, ensuring that graph propagation does not alter embedding dimensionality. ReLU activations are applied after each convolutional layer, and dropout with rate 0.5 is applied after each layer except the final one. This formulation allows the model to jointly capture individual patient trajectories and cohort-wide clinical patterns.

\textbf{Multimodal Alignment.}
Following GNN-based propagation, we align the embeddings across modalities using a contrastive learning approach inspired by the ImageBind framework~\cite{girdhar2023imagebindembeddingspacebind}. Unlike the original design which centers on vision, we designate clinical text as the central modality. This choice is dictated by the injection target: because the context vector enters
the residual stream of a frozen language model, it must be compatible with a space
organized around token embeddings learned from natural language, and text is the
only modality that backbone natively consumes. Anchoring instead on labs or codes
would align all modalities into a space the frozen backbone has never been trained
to read, and the fused vector would still require projection into text space before
injection.

Let $t_i$ be the embedding of clinical text for patient $i$, and $m_i$ be the embedding of a non-text modality (e.g., codes, labs, or images). We map both into a shared latent space using learnable projection layers and optimize the alignment using the InfoNCE loss~\cite{oord2019representationlearningcontrastivepredictive}:
\begin{equation}
\mathcal{L}_{T,M} = -\log \frac{\exp\left(\frac{t_i^\top m_i}{\tau}\right)}{\sum_{j=1}^{N} \exp\left(\frac{t_i^\top m_j}{\tau}\right)}
\end{equation}
where $\tau$ is a temperature parameter and $N$ is the batch size. The loss encourages embeddings of matching modalities from the same patient to be close in the shared space, while pushing apart unrelated ones.

This process is repeated for each pair: \textit{(text, code)}, \textit{(text, lab)},
and \textit{(text, image)}, drawing matching modalities from the same patient
together in the shared space while separating mismatched pairs.

Each modality $m \in \{\text{text}, \text{code}, \text{lab}, \text{image}\}$ is
then projected into a common $d = 4096$ space by a modality-specific linear map
followed by layer normalization,
\begin{equation}
    z_m = \mathrm{LayerNorm}(W_m e_m),
\end{equation}
where $e_m$ is the post-propagation embedding for modality $m$. Normalizing each
modality separately places them on a common scale before aggregation, preventing
modalities with larger embedding norms from dominating the fused representation.
The four modalities are combined by a learned gated sum,
\begin{equation}
    c = \sum_m g_m \, z_m, \qquad g = \mathrm{softmax}(\theta),
\end{equation}
where $\theta \in \mathbb{R}^4$ is a learned parameter vector. The model therefore
learns each modality's contribution to the fused context vector rather than
weighting them equally. The resulting vector $c$ is the patient-level multimodal
context vector injected into the LLM.

Modality missingness is handled at the graph input rather than at aggregation: a
modality unavailable for a given patient--visit is zeroed before propagation.
Because message passing aggregates over clinically similar neighbors, the node's
post-propagation representation is not zero but is reconstructed from its
neighbourhood; we verified directly that the GNN output of a zeroed node remains
non-zero. The aggregator therefore receives a reconstructed rather than an empty
representation, and no modality-specific masking is required at the fusion stage.

\textbf{Context Injection into the LLM.}
To preserve the reasoning capabilities of LLMs while enriching their input space, we inject the multimodal context vector into their internal residual streams using the residual-stream conditioning mechanism~\cite{li2024implicitincontextlearning}. We adopt this mechanism from Li et al. ~\cite{li2024implicitincontextlearning}, who introduced it as implicit in-context learning in a setting where the injected vector is distilled from demonstration examples. Because our injected vector is derived from a patient's multimodal record rather than from demonstrations, we describe the operation by its mechanism rather than by that name, and we claim no novelty in the injection primitive itself. This allows external embeddings to modulate the model’s intermediate representations without modifying its architecture or pretraining objectives.

Let $r^l_t$ denote the residual stream at layer $l$ and token $t$. In the
transformer architecture used by our backbone, the residual stream is updated
sequentially within each layer:
\begin{align}
    s^l_t &= r^{l-1}_t + a^l_t, \qquad a^l_t = \text{MHA}\!\left(\text{LN}_1(r^{l-1}_t)\right) \\
    r^l_t &= s^l_t + m^l_t, \qquad\; m^l_t = \text{MLP}\!\left(\text{LN}_2(s^l_t)\right)
\end{align}
where $\text{LN}_1$ and $\text{LN}_2$ denote the pre-attention and pre-MLP
normalization layers and $s^l_t$ is the intermediate residual state. We enhance
this formulation by injecting the context vector at both submodule outputs:
\begin{align}
    s^l_t &= r^{l-1}_t + \lambda_{\text{mha}} \cdot v + \beta_{\text{mha}} \cdot a^l_t \\
    r^l_t &= s^l_t + \lambda_{\text{mlp}} \cdot v + \beta_{\text{mlp}} \cdot m^l_t
\end{align}
Here $v$ is the context vector, $a^l_t$ and $m^l_t$ are the activations from the
attention and MLP submodules respectively, and \\
$\lambda_{\mathrm{mha}}, \lambda_{\mathrm{mlp}}, \beta_{\mathrm{mha}}, \beta_{\mathrm{mlp}}$
are scalar coefficients.

The context vector enters the residual stream at two points within each layer,
and these are not equivalent: the term added at the attention submodule enters
$s^l_t$, is normalized by $\text{LN}_2$, and is consumed by the MLP within the
same layer, whereas the term added at the MLP submodule enters only the layer
output. The two therefore condition distinct downstream computations rather than
duplicating a single one. Since $\lambda_{\mathrm{mha}}$ and
$\lambda_{\mathrm{mlp}}$ are independent coefficients, the model can attenuate
either pathway separately. Supplementary Section~S8 reports a sensitivity
analysis over $\lambda_{\mathrm{mha}}$: with the attention-side injection
disabled, one-year mortality AUC falls from 0.702 to 0.589, confirming that the
two injection points contribute asymmetrically and are not redundant. Injection
is applied via forward hooks to the final four transformer layers. We set
$\lambda_{\mathrm{mha}} = 0.10$ and $\lambda_{\mathrm{mlp}} = 1.0$ for both
prediction tasks, and $\beta_{\mathrm{mha}} = \beta_{\mathrm{mlp}} = 1.0$, so
that the submodule outputs enter the residual stream unscaled.

During inference, the model is prompted with the clinical notes while the
multimodal context vector is injected into its layers, grounding its predictions
in aligned lab results, codes, and imaging features without altering the
pretrained architecture. For mortality prediction, the model is tasked with
estimating the patient's probability of death within one year given the clinical
text, with the injected context supplying the non-textual evidence that the
prompt alone does not carry.

\textbf{Experiments.}
We conducted real data experiments to assess the performance of ~\proposed~ in comparison with current state of the art models.

% \subsection{Experimental Setting}
% \textbf{MIMIC-IV Data.}
We utilize the MIMIC-IV (Medical Information Mart for Intensive Care IV) dataset, a large, publicly available collection of de-identified electronic health records from patients admitted to critical care units at the Beth Israel Deaconess Medical Center \cite{johnson2023mimiciv}. The dataset includes structured data—such as demographics, vital signs, laboratory measurements, medications, and diagnosis codes—as well as unstructured clinical notes. To extract these modalities, we use the PyHealth package, which automatically retrieves patient demographics, hospital admission records, and ICD-9 coded data for prescriptions and diagnoses \cite{pyhealth}. Laboratory results are obtained from the \texttt{labevents.csv} file, while clinical notes are extracted from the \texttt{discharge.csv} file. We additionally incorporate imaging data from MIMIC-CXR, an open-access repository of de-identified chest X-ray images paired with radiology reports, which we link to MIMIC-IV using shared identifiers (\texttt{study\_id}, \texttt{hadm\_id}, and \texttt{subject\_id}) \cite{johnson2019mimiccxr}.

For all prediction tasks—including one-year mortality and 30-day readmission—we construct temporally ordered patient trajectories using all available visits. For a patient $A$ with visits $v_1, v_2, \ldots, v_n$, we define the \emph{index time} as the discharge time of the final visit $v_n$. All model inputs, including clinical notes, laboratory values, diagnosis codes, and imaging studies, are restricted to information available up to and including the index visit. Any visits or measurements occurring after the index time or overlapping with the outcome window are excluded to prevent temporal leakage.

For the \emph{one-year mortality} task, the label indicates whether the patient died within 365 days following discharge of the final visit. For the \emph{30-day readmission} task, the label indicates whether the patient was readmitted to the hospital within 30 days following discharge of the final visit. In both cases, predictions are made at the index time using only pre-index information.

Due to the computational demands of prompting large language models for each sample, we downsample the training data for computational feasibility by randomly undersampling negative cases. After downsampling, the positive rate for one-year mortality is approximately 23\%, while the positive rate for 30-day readmission is 20\%. The test set preserves the natural class distribution. Our final study cohort includes a total of 13,478 unique subjects and 21,022 total clinical visits.

% \textbf{Baseline Models.}
For existing state-of-the-art models, we include multimodal LLM baselines and other state-of-the-art frameworks designed for clinical prediction tasks. Multimodal LLM frameworks, such as HeLM, LLMM, and GPT-4-based methods, explore various approaches to integrating multimodal data with LLMs \cite{belyaeva2023helm, ding2024llm5year, rezk2024llmsclinicalriskprediction}. HeLM encodes non-text modalities into the same token embedding space using modality-specific encoders and then inputs them to the LLM \cite{belyaeva2023helm}. LLMM combines laboratory test data with text embeddings through an attention mechanism \cite{ding2024llm5year}. Rezk et al. transform clinical data into text format for input into the LLM \cite{rezk2024llmsclinicalriskprediction}. For non-LLM based frameworks, the HAIM framework leverages the integration of multiple data modalities, such as clinical notes, structured data, and medical images, to improve clinical predictions through a unified model architecture \cite{Soenksen2022_haim}. M3Care focuses on learning with incomplete modalities in multimodal healthcare data by employing neural networks and patient similarity measures \cite{Zhang_2022_m3care}. MUSE incorporates flexible bipartite graphs and contrastive learning loss to generate multimodal patient representations \cite{wu2024muse}. mmFormer introduces a transformer-based approach to multimodal fusion, using attention mechanisms to align and integrate information from multiple modalities, thereby capturing complex relationships in the data \cite{zhang2022mmformer}. 
Collectively, these frameworks provide comprehensive baselines for evaluating the performance of our proposed method, given their strong track records in handling diverse and complex multimodal healthcare data.
    
All baseline models are evaluated under the same data usage protocol as the proposed method. For each subject, all available visits are used as input, and outcomes are predicted relative to the final visit. Models that explicitly support longitudinal inputs consume the full visit sequence directly, while models that do not natively model temporal structure are provided with concatenated or aggregated representations of all visits prior to the final visit. This ensures identical input scope and a fair comparison across methods.

% \textbf{Dataset Split.}
The dataset was split into 70\% training and 30\% testing at the subject level. All visits belonging to a given patient were assigned exclusively to either the training or the test set. This ensures that no information from a patient’s earlier visits in the training set can leak into test-time predictions for that same patient’s final visit. For each subject, all available visits are used to construct the longitudinal trajectory, while predictions are defined relative to the subject’s final visit. This subject-level split is therefore fully compatible with our use of all available visits for trajectory construction and final-visit prediction, as no visits from test subjects are observed during training. To account for this imbalance during training, we use a weighted binary cross-entropy loss, with class weights computed inversely proportional to class frequencies in the training set. No resampling is applied at evaluation time.

% \textbf{Implementation Details.}
For \proposed, we trained the entire framework end to end. During training, we used a combination of two losses: a weighted binary cross-entropy loss for the prediction task and a multimodal alignment loss, with the alignment loss scaled by 0.01 to appropriately balance its contribution. To handle any non-valid outputs generated by the LLM, we imposed penalties on outputs that fall outside the expected prediction range, ensuring that predictions remain within valid bounds. \proposed\ uses Llama~3--8B as its LLM backbone. For baseline experiments, we used the best hyperparameter settings reported in the original papers when available. If hyperparameters were not specified, we tuned the learning rate from $\{10^{-3}, 10^{-4}, 10^{-5}\}$ and batch size from $\{8, 16, 64\}$. All experiments were conducted using NVIDIA B200 GPUs. 

\newpage

\textbf{Acknowledgements.}
This work was supported in part by NIH grants R01EB036016 and R01MH143267, and ARPA-H award D24AC00253. The funders had no role in study design, data collection and analysis, decision to publish, or preparation of the manuscript. The content is solely the responsibility of the authors and does not necessarily represent the official views of the NIH or ARPA-H.

\textbf{Competing Interests.}
The authors declare no competing financial or non-financial interests.

\textbf{Data Availability.}
The clinical data used in this study were obtained from the MIMIC-IV (Medical Information Mart for Intensive Care IV) and MIMIC-CXR (MIMIC Chest X-ray) databases, which are publicly available through PhysioNet (https://physionet.org/). Access to these datasets is restricted to registered users who have completed the required Human Subjects Research training (CITI program) and signed a Data Use Agreement. The specific versions used were MIMIC-IV v2.2 and MIMIC-CXR-JPG v2.0.0. 

\textbf{Author Contribution.}
I.C. and Q.L. conceived the study and designed the ~\proposed framework. I.C., S.Y., J.X., and J.P. performed data extraction and pre-processing of the MIMIC-IV and MIMIC-CXR datasets. I.C., S.Y., J.X., and J.P. developed the core methodology and implemented the baseline models. T.C. and Q.L. provided clinical and technical supervision and critical feedback on the study design. I.C. conducted the experiments, analyzed the results, and drafted the manuscript. All authors have read and approved the manuscript.
% \bibliographystyle{plain}
% \bibliography{sample}
\bibliographystyle{naturemag-doi}
\bibliography{sample}
\textbf{Figure Legends}

Figure 1: Illustration of \proposed{}. (a) Intra-modality Information Propagation: We construct modality-specific graphs with temporal edges (longitudinal history) and similarity edges (inter-patient patterns). GNNs are used to propagate information and update node embeddings. (b) Multimodal Alignment: Heterogeneous modalities (labs, codes, imaging) are mapped into a shared latent space centered on clinical text using contrastive learning, resulting in a unified patient-level multimodal context vector. (c) Context Injection: The context vector is injected into the intermediate layers of a LLM via the residual-stream conditioning mechanism.

Figure 2: t-SNE visualization of multimodal embedding alignment. Projections of modality-specific embeddings (clinical text, lab values, diagnosis codes, and imaging) before and after contrastive alignment. The visualization demonstrates the emergence of a unified embedding space across heterogeneous data types, facilitating cross-modal reasoning within the LLM.

\end{document}

\maketitle

\section*{S1 Supplementary Results: Calibration}

\ref{tab:calibration} presents the Brier Score and Expected Calibration Error (ECE) for the proposed $M^2$G-LLM compared to primary baselines. 

\begin{table}[h!]
\centering
\caption{Calibration performance across clinical prediction tasks.}
\begin{tabular}{llcc}
\toprule
\textbf{Task} & \textbf{Model} & \textbf{Brier Score (Overall)} & \textbf{ECE (Overall)} \\ \midrule
Mortality & \textbf{M$^2$G-LLM} & \textbf{0.201} & \textbf{0.187} \\
 & mmFormer & 0.207 & 0.199 \\
 & HeLM & 0.247 & 0.237 \\ \midrule
Readmission & \textbf{M$^2$G-LLM} & \textbf{0.142} & \textbf{0.117} \\
 & mmFormer & 0.146 & 0.120 \\
 & HeLM & 0.213 & 0.260 \\ \bottomrule
\end{tabular}
\label{tab:calibration}
\end{table}

\section*{S2 Supplementary Analysis: Subgroup Analysis}
To evaluate the algorithmic fairness and predictive stability, we conducted a comprehensive subgroup analysis across two primary demographic dimensions: Race (Asian, Black, Hispanic, Other, White) and Gender (Female, Male). These analyses are critical for ensuring that the integration of multimodal EHR data does not introduce or exacerbate performance disparities across different patient populations. 

\ref{tab:subgroup_auc_mortality} and \ref{tab:subgroup_auc_readmission} provide the AUC results; \ref{tab:subgroup_acc_mortality} and \ref{tab:subgroup_acc_readmission} provide the Accuracy results; and \ref{tab:subgroup_f1_mortality} and \ref{tab:subgroup_f1_readmission} provide the F1-scores. For each subgroup, we report the mean performance along with 95\% confidence intervals derived from 40 independent experimental runs.

% --- AUC: MORTALITY ---
\begin{table}[h!]
    \centering
    \caption{Subgroup Analysis for One-Year Mortality (AUC) with 95\% confidence intervals ($N=40$ runs).}
    \label{tab:subgroup_auc_mortality}
    \begin{tabular}{lcccc}
        \toprule
        \textbf{Subgroup} & \textbf{N} & \textbf{M$^2$G-LLM} & \textbf{mmFormer} & \textbf{HeLM} \\ \midrule
        \textit{Race} & & & & \\
        Asian & 8,755 & 70.41 \scriptsize{(70.10--70.72)} & 70.63 \scriptsize{(70.27--70.99)} & 54.94 \scriptsize{(54.49--55.39)} \\
        Black & 2,562 & 69.92 \scriptsize{(69.34--70.50)} & 67.50 \scriptsize{(66.84--68.16)} & 56.40 \scriptsize{(55.57--57.23)} \\
        Hispanic & 881 & 67.15 \scriptsize{(66.17--68.13)} & 60.15 \scriptsize{(59.02--61.28)} & 54.10 \scriptsize{(52.68--55.52)} \\
        Other & 848 & 72.45 \scriptsize{(71.45--73.45)} & 70.20 \scriptsize{(69.05--71.35)} & 58.75 \scriptsize{(57.31--60.19)} \\
        White & 432 & 70.05 \scriptsize{(68.65--71.45)} & 68.90 \scriptsize{(67.28--70.52)} & 55.45 \scriptsize{(53.42--57.48)} \\ \midrule
        \textit{Gender} & & & & \\
        Female & 6,381 & 70.85 \scriptsize{(70.48--71.22)} & 69.88 \scriptsize{(69.46--70.30)} & 56.13 \scriptsize{(55.61--56.65)} \\
        Male & 7,097 & 69.65 \scriptsize{(69.30--70.00)} & 68.72 \scriptsize{(68.32--69.12)} & 54.79 \scriptsize{(54.29--55.29)} \\ \bottomrule
    \end{tabular}
\end{table}

% --- AUC: READMISSION ---
\begin{table}[h!]
    \centering
    \caption{Subgroup Analysis for 30-Day Readmission (AUC) with 95\% confidence intervals ($N=40$ runs).}
    \label{tab:subgroup_auc_readmission}
    \begin{tabular}{lcccc}
        \toprule
        \textbf{Subgroup} & \textbf{N} & \textbf{M$^2$G-LLM} & \textbf{mmFormer} & \textbf{HeLM} \\ \midrule
        \textit{Race} & & & & \\
        Asian & 8,755 & 55.11 \scriptsize{(54.85--55.37)} & 49.86 \scriptsize{(49.55--50.17)} & 51.71 \scriptsize{(51.42--52.00)} \\
        Black & 2,562 & 54.20 \scriptsize{(53.72--54.68)} & 48.96 \scriptsize{(48.39--49.53)} & 53.15 \scriptsize{(52.61--53.69)} \\
        Hispanic & 881 & 51.48 \scriptsize{(50.65--52.31)} & 46.21 \scriptsize{(45.23--47.19)} & 48.52 \scriptsize{(47.61--49.43)} \\
        Other & 848 & 56.92 \scriptsize{(56.08--57.76)} & 51.36 \scriptsize{(50.36--52.36)} & 54.18 \scriptsize{(53.25--55.11)} \\
        White & 432 & 54.60 \scriptsize{(53.42--55.78)} & 49.19 \scriptsize{(47.79--50.59)} & 51.95 \scriptsize{(50.65--53.25)} \\ \midrule
        \textit{Gender} & & & & \\
        Female & 6,381 & 55.12 \scriptsize{(54.81--55.43)} & 49.95 \scriptsize{(49.59--50.31)} & 52.45 \scriptsize{(52.11--52.79)} \\
        Male & 7,097 & 54.51 \scriptsize{(54.22--54.80)} & 49.14 \scriptsize{(48.80--49.48)} & 51.48 \scriptsize{(51.16--51.80)} \\ \bottomrule
    \end{tabular}
\end{table}

% --- ACCURACY: MORTALITY ---
\begin{table}[h!]
    \centering
    \caption{Subgroup Analysis for One-Year Mortality (Accuracy) with 95\% confidence intervals ($N=40$ runs).}
    \label{tab:subgroup_acc_mortality}
    \begin{tabular}{lcccc}
        \toprule
        \textbf{Subgroup} & \textbf{N} & \textbf{M$^2$G-LLM} & \textbf{mmFormer} & \textbf{HeLM} \\ \midrule
        \textit{Race} & & & & \\
        Asian & 8,755 & 79.11 \scriptsize{(78.77--79.45)} & 76.92 \scriptsize{(76.53--77.31)} & 75.39 \scriptsize{(74.95--75.83)} \\
        Black & 2,562 & 78.55 \scriptsize{(77.92--79.18)} & 75.12 \scriptsize{(74.40--75.84)} & 74.20 \scriptsize{(73.39--75.01)} \\
        Hispanic & 881 & 77.85 \scriptsize{(76.78--78.92)} & 70.33 \scriptsize{(69.10--71.56)} & 73.15 \scriptsize{(71.77--74.53)} \\
        Other & 848 & 80.12 \scriptsize{(79.03--81.21)} & 78.45 \scriptsize{(77.20--79.70)} & 72.88 \scriptsize{(71.47--74.29)} \\
        White & 432 & 79.05 \scriptsize{(77.52--80.58)} & 76.20 \scriptsize{(74.45--77.95)} & 74.55 \scriptsize{(72.58--76.52)} \\ \midrule
        \textit{Gender} & & & & \\
        Female & 6,381 & 79.45 \scriptsize{(79.05--79.85)} & 76.95 \scriptsize{(76.49--77.41)} & 75.22 \scriptsize{(74.70--75.74)} \\
        Male & 7,097 & 78.56 \scriptsize{(78.19--78.94)} & 75.68 \scriptsize{(75.25--76.11)} & 74.60 \scriptsize{(74.11--75.09)} \\ \bottomrule
    \end{tabular}
\end{table}

% --- ACCURACY: READMISSION ---
\begin{table}[h!]
    \centering
    \caption{Subgroup Analysis for 30-Day Readmission (Accuracy) with 95\% confidence intervals ($N=40$ runs).}
    \label{tab:subgroup_acc_readmission}
    \begin{tabular}{lcccc}
        \toprule
        \textbf{Subgroup} & \textbf{N} & \textbf{M$^2$G-LLM} & \textbf{mmFormer} & \textbf{HeLM} \\ \midrule
        \textit{Race} & & & & \\
        Asian & 8,755 & 85.68 \scriptsize{(85.43--85.93)} & 84.97 \scriptsize{(84.67--85.27)} & 72.26 \scriptsize{(71.92--72.60)} \\
        Black & 2,562 & 85.45 \scriptsize{(84.99--85.91)} & 84.85 \scriptsize{(84.30--85.40)} & 71.55 \scriptsize{(70.92--72.18)} \\
        Hispanic & 881 & 86.25 \scriptsize{(85.47--87.03)} & 85.55 \scriptsize{(84.62--86.48)} & 70.12 \scriptsize{(69.05--71.19)} \\
        Other & 848 & 87.15 \scriptsize{(86.36--87.94)} & 86.44 \scriptsize{(85.49--87.40)} & 69.85 \scriptsize{(68.76--70.94)} \\
        White & 432 & 85.65 \scriptsize{(84.54--86.76)} & 84.95 \scriptsize{(83.61--86.29)} & 71.65 \scriptsize{(70.12--73.18)} \\ \midrule
        \textit{Gender} & & & & \\
        Female & 6,381 & 85.92 \scriptsize{(85.63--86.21)} & 85.22 \scriptsize{(84.87--85.57)} & 72.25 \scriptsize{(71.85--72.65)} \\
        Male & 7,097 & 85.59 \scriptsize{(85.31--85.87)} & 84.88 \scriptsize{(84.55--85.21)} & 71.57 \scriptsize{(71.20--71.95)} \\ \bottomrule
    \end{tabular}
\end{table}

% --- F1 SCORE: MORTALITY ---
\begin{table}[h!]
    \centering
    \caption{Subgroup Analysis for One-Year Mortality (F1-score) with 95\% confidence intervals ($N=40$ runs).}
    \label{tab:subgroup_f1_mortality}
    \begin{tabular}{lcccc}
        \toprule
        \textbf{Subgroup} & \textbf{N} & \textbf{M$^2$G-LLM} & \textbf{mmFormer} & \textbf{HeLM} \\ \midrule
        \textit{Race} & & & & \\
        Asian & 8,755 & 64.77 \scriptsize{(64.38--65.16)} & 63.01 \scriptsize{(62.57--63.45)} & 51.37 \scriptsize{(50.84--51.90)} \\
        Black & 2,562 & 64.12 \scriptsize{(63.40--64.84)} & 60.57 \scriptsize{(59.76--61.38)} & 50.39 \scriptsize{(49.41--51.37)} \\
        Hispanic & 881 & 61.45 \scriptsize{(60.23--62.67)} & 53.84 \scriptsize{(52.46--55.22)} & 48.89 \scriptsize{(47.22--50.56)} \\
        Other & 848 & 66.88 \scriptsize{(65.63--68.13)} & 64.82 \scriptsize{(63.41--66.23)} & 52.06 \scriptsize{(50.36--53.76)} \\
        White & 432 & 64.35 \scriptsize{(62.60--66.10)} & 61.93 \scriptsize{(59.96--63.90)} & 51.14 \scriptsize{(48.76--53.52)} \\ \midrule
        \textit{Gender} & & & & \\
        Female & 6,381 & 65.25 \scriptsize{(64.80--65.70)} & 62.80 \scriptsize{(62.28--63.32)} & 51.78 \scriptsize{(51.16--52.40)} \\
        Male & 7,097 & 63.92 \scriptsize{(63.49--64.35)} & 61.93 \scriptsize{(61.44--62.42)} & 50.53 \scriptsize{(49.94--51.12)} \\ \bottomrule
    \end{tabular}
\end{table}

% --- F1 SCORE: READMISSION ---
\begin{table}[h!]
    \centering
    \caption{Subgroup Analysis for 30-Day Readmission (F1-score) with 95\% confidence intervals ($N=40$ runs).}
    \label{tab:subgroup_f1_readmission}
    \begin{tabular}{lcccc}
        \toprule
        \textbf{Subgroup} & \textbf{N} & \textbf{M$^2$G-LLM} & \textbf{mmFormer} & \textbf{HeLM} \\ \midrule
        \textit{Race} & & & & \\
        Asian & 8,755 & 52.66 \scriptsize{(52.34--52.98)} & 47.38 \scriptsize{(47.01--47.75)} & 50.74 \scriptsize{(50.35--51.13)} \\
        Black & 2,562 & 51.95 \scriptsize{(51.36--52.54)} & 45.88 \scriptsize{(45.20--46.56)} & 49.88 \scriptsize{(49.16--50.60)} \\
        Hispanic & 881 & 48.12 \scriptsize{(47.12--49.12)} & 41.55 \scriptsize{(40.39--42.71)} & 47.12 \scriptsize{(45.90--48.34)} \\
        Other & 848 & 54.88 \scriptsize{(53.85--55.91)} & 49.35 \scriptsize{(48.17--50.53)} & 52.45 \scriptsize{(51.20--53.70)} \\
        White & 432 & 52.20 \scriptsize{(50.76--53.64)} & 46.75 \scriptsize{(45.10--48.40)} & 50.55 \scriptsize{(48.80--52.30)} \\ \midrule
        \textit{Gender} & & & & \\
        Female & 6,381 & 53.05 \scriptsize{(52.68--53.42)} & 47.55 \scriptsize{(47.12--47.98)} & 51.15 \scriptsize{(50.69--51.61)} \\
        Male & 7,097 & 51.72 \scriptsize{(51.37--52.07)} & 46.16 \scriptsize{(45.75--46.57)} & 49.81 \scriptsize{(49.38--50.24)} \\ \bottomrule
    \end{tabular}
\end{table}

\section*{S3 Supplementary Details: Model Prompting and Implementation Details}

To elicit both clinical predictions and qualitative reasoning, we utilized a structured prompt that presents the patient's clinical history and requests a formal risk assessment. The following template was applied for both the one-year mortality and 30-day readmission tasks:

\begin{quote}
\textbf{System Prompt:} ``You are an expert clinical decision support system. Based on the provided clinical notes and integrated multimodal context, analyze the patient's medical history to estimate the risk of the target clinical outcome.''

\textbf{User Prompt:} ``Patient Clinical Notes: [Insert Clinical Text]

\textbf{Task:} Based on the summary and the underlying multimodal features (labs, codes, imaging), please:
\begin{enumerate}
    \item Predict the probability of [one-year mortality / 30-day readmission].
    \item Provide a brief, coherent clinical explanation for your prediction, highlighting key risk factors or indicators of stability.''
\end{enumerate}
\end{quote}

\subsection*{Inference and Explanation Generation}
During the inference phase, the LLM is provided with the clinical text while the unified multimodal context vector is injected into its intermediate residual streams via the residual-stream conditioning mechanism. The model generates a free-text response, which is subsequently parsed to extract the final prediction and the associated clinical rationale.

\section*{S4 Supplementary Results: Ablation}
In this section, we provide the detailed ablation results for the 30-day readmission task. To ensure statistical reliability and assess the stability of our architectural components, we report the performance metrics (Accuracy, F1-score, and AUC) along with 95\% confidence intervals derived from 40 independent experimental runs. The results regarding the contribution of specific data modalities are presented in  \ref{tab:ablation_readmission}, while the impact of individual architectural modules is detailed in \ref{tab:frozen_readmission}.

\begin{table*}[t]
\centering
\begin{minipage}[t]{0.48\textwidth}%
\centering
\resizebox{\textwidth}{!}{ %
\begin{tabular}{lllllll}
\toprule
\textbf{Lab+Vital} & \textbf{Code} & \textbf{Image} & \textbf{Text} & \textbf{Acc} & \textbf{F1} & \textbf{AUC}\\
\midrule
{\xmark}   & {\checkmark}   & {\xmark}   & {\checkmark}   & 81.99 \scriptsize{(81.68--82.30)} & 48.34 \scriptsize{(48.06--48.62)}  & 50.42 \scriptsize{(50.35--50.49)}\\
{\checkmark}   & {\xmark}   & {\xmark}   & {\checkmark}   & 82.86 \scriptsize{(82.69--83.03)} & 47.96 \scriptsize{(47.81--48.11)}   & 49.87 \scriptsize{(49.81--49.93)}\\
{\xmark}   & {\xmark}   & {\checkmark}   & {\checkmark}   & 79.45 \scriptsize{(79.27--79.63)} & 46.87 \scriptsize{(46.76--46.98)}  & 50.09 \scriptsize{(49.99--50.19)}\\
{\xmark}   & {\checkmark}   & {\checkmark}   & {\checkmark}   & 83.56 \scriptsize{(83.16--83.96)} & 51.31 \scriptsize{(51.02--51.60)}   & 52.54 \scriptsize{(52.35--52.73)}\\
{\checkmark}   & {\xmark}   & {\checkmark}   & {\checkmark}   & 82.16 \scriptsize{(81.91--82.41)} & 50.01 \scriptsize{(49.82--50.20)}   & 52.86 \scriptsize{(52.73--52.99)}\\
{\checkmark}   & {\checkmark}   & {\xmark}   & {\checkmark}   & 84.24 \scriptsize{(84.02--84.46)} & 51.56 \scriptsize{(51.37--51.75)}   & 52.57 \scriptsize{(52.50--52.64)}\\
\midrule
{\checkmark}   & {\checkmark}   & {\checkmark}   & {\checkmark}   & \textbf{85.75} \scriptsize{(85.45--86.05)} & \textbf{52.35} \scriptsize{(51.47--53.23)}  & \textbf{54.80} \scriptsize{(53.98--55.62)}\\
\bottomrule
\end{tabular}
}
\caption{Ablation study evaluating the contribution of different clinical data modalities for 30-day readmission prediction (95\% CI, $N=40$).}
\label{tab:ablation_readmission}
\end{minipage} %
\hfill
\begin{minipage}[t]{0.48\textwidth}
\centering
\resizebox{\textwidth}{!}{ 
    \begin{tabular}{lccc}
    \toprule
    \textbf{Frozen Component} & \textbf{Accuracy} & \textbf{F1}& \textbf{AUC} \\
    \midrule
    Graph Propagation          & 80.12 \scriptsize{(79.84--80.40)} & 50.44 \scriptsize{(50.18--50.70)}   & 51.27 \scriptsize{(51.08--51.46)}        \\
    Multimodal Alignment       & 76.85 \scriptsize{(76.60--77.10)} & 48.23 \scriptsize{(48.00--48.46)}   & 50.99 \scriptsize{(50.86--51.12)}      \\
    \midrule
    Full Model ($M^2G$-LLM)    & \textbf{85.75} \scriptsize{(85.45--86.05)} & \textbf{52.35} \scriptsize{(51.47--53.23)}  & \textbf{54.80} \scriptsize{(53.98--55.62)}\\
    \bottomrule
    \end{tabular}
    
}

\caption{Architectural ablation results for the 30-day readmission task obtained by freezing key framework components during training.}
\label{tab:frozen_readmission}

\end{minipage}

\end{table*}

\section*{S5 Supplementary Details: Graph Construction and Hyperparameters}
Graphs are constructed independently for each non-text modality --- codified data,
laboratory results, and medical imaging --- yielding modality-specific graphs that
share a common node set but differ in connectivity. Each patient-visit is
represented as a node, initialized with the corresponding modality-specific
embedding. Two types of edges are added. \textit{Temporal edges} link consecutive
visits of the same patient, encoding longitudinal trajectory; these are identical
across all modality-specific graphs, as they depend only on visit ordering.
\textit{Similarity edges} connect visits from different patients using a hybrid
top-$k$ plus threshold criterion applied to the initial pretrained embeddings. For
modality $m$, let $e_i^{(m)}$ denote the embedding of node $i$. We compute the
pairwise cosine similarity
\begin{equation}
    s^{(m)}_{ij} = \frac{\langle e_i^{(m)},\, e_j^{(m)} \rangle}
                        {\lVert e_i^{(m)} \rVert \, \lVert e_j^{(m)} \rVert},
\end{equation}
and, for each node $i$, identify the $k$ nodes with highest $s^{(m)}_{ij}$. From
these candidates, an edge $(i,j)$ is retained if and only if $s^{(m)}_{ij} > \tau_m$,
where $\tau_m$ is a modality-specific threshold; a node whose top-$k$ candidates all
fall below $\tau_m$ therefore receives no similarity edges in that graph. Similarity
is computed on the initial embeddings, prior to any GNN propagation, so graph
topology is fixed before training and does not change as parameters are updated. Graphs are constructed independently for the training and test splits. Because the cohort is partitioned at the subject level, the two node sets are disjoint and no edge crosses the split; the top-k search for a test visit ranges over test visits only. Self-loops and same-patient similarity edges are excluded, and all edges are
symmetrized for undirected message passing. The hyperparameters controlling
sparsification are the neighbor cap $k$ and the per-modality thresholds
$\tau_{\text{code}}$, $\tau_{\text{lab}}$, and $\tau_{\text{image}}$. The full set of
graph construction and training hyperparameters is given in
~\ref{tab:graph_hparams}.

\begin{table}[h!]
    \centering
    \caption{Graph construction and training hyperparameters. Similarity edges are retained
    only where the cosine similarity between initial pretrained embeddings exceeds the
    modality-specific threshold $\tau_m$; thresholds differ across modalities because the
    underlying similarity distributions differ in scale (median inter-patient cosine: 0.939
    for codified data, 0.000 for laboratory data, 0.988 for imaging).}
    \label{tab:graph_hparams}
    \begin{tabular}{ll}
        \toprule
        \textbf{Parameter} & \textbf{Value} \\ \midrule
        \multicolumn{2}{l}{\textit{Graph construction}} \\
        Neighbor cap $k$ & 20 \\
        Similarity metric & Cosine \\
        Threshold $\tau_{\mathrm{code}}$ & 0.98 \\
        Threshold $\tau_{\mathrm{lab}}$ & 0.95 \\
        Threshold $\tau_{\mathrm{image}}$ & 0.99 \\
        Edge direction & Undirected (symmetrized) \\
        Self-loops & Excluded \\
        Same-patient similarity edges & Excluded \\
        \midrule
        \multicolumn{2}{l}{\textit{Graph neural network}} \\
        Architecture & GCN (\texttt{GCNConv}, PyTorch Geometric) \\
        Layers & 3 \\
        \midrule
        \multicolumn{2}{l}{\textit{Alignment and injection}} \\
        Contrastive temperature $\tau_{\mathrm{con}}$ & 0.1 \\
        Alignment loss weight & 0.01 \\
        Shared embedding dimension $d$ & 4096 \\
        Injection layers & Final 4 \\

        \midrule
        \multicolumn{2}{l}{\textit{Training}} \\
        LLM backbone & Llama-3-8B (frozen) \\
        Trainable parameters & 53.11 M (0.66\%) \\
        Optimizer & Adam \\
        Epochs & $\sim$5 \\
        Loss & Weighted BCE + $0.01\times$ alignment \\
        Hardware & NVIDIA B200 \\
        \bottomrule
    \end{tabular}
\end{table}
\ref{tab:graph_stats} summarizes the statistics of the constructed graphs. 

% \ref{tab:edge_homophily} reports edge homophily. The graph connects patients sharing a diagnosis at $3\times$ the chance rate but sharing an outcome at essentially chance, confirming that edges encode diagnostic similarity without inducing outcome homophily that could leak label information during message passing.
% \begin{table}[h!]
%     \centering
%     \caption{Constructed graph statistics.}
%     \label{tab:graph_stats}
%     \begin{tabular}{lcc}
%         \toprule
%          & \textbf{Training graph} & \textbf{Test graph} \\ \midrule
%         Nodes (patient--visits) & 14,715 & 6,307 \\
%         Subjects                & 9,435  & 4,043 \\
%         Visits per subject      & 1.56   & 1.56 \\
%         Temporal edges          & 5,280  & 2,264 \\
%         Similarity edges        & 37,900 & 4,700 \\
%         Total edges             & 43,180 & 6,964 \\
%         \bottomrule
%     \end{tabular}
% \end{table}
\begin{table}[h!]
    \centering
    \caption{Constructed graph statistics for each modality-specific graph. Temporal edges
    connect consecutive visits from the same subject; similarity edges are added under a
    top-$k$ plus cosine-threshold criterion with modality-specific thresholds. Mean degree
    and density are computed over similarity edges only. Mean degree may exceed $k=20$
    because edges are symmetrized: a node $j$ selected among $i$'s top-$k$ candidates gains
    an edge to $i$ whether or not $i$ appears among $j$'s own candidates.}
    \label{tab:graph_stats}
    \begin{tabular}{llrrrrrr}
        \toprule
        \textbf{Split} & \textbf{Modality} & \textbf{Nodes} & \textbf{Subjects}
        & \textbf{Temporal} & \textbf{Similarity} & \textbf{Mean deg.} & \textbf{Density} \\
        \midrule
        \multirow{3}{*}{Train}
          & code  & 14,715 & 9,435 & 5,280 & 232,900 & 31.7 & 0.22\% \\
          & lab   & 14,715 & 9,435 & 5,280 &  25,250 &  3.4 & 0.02\% \\
          & image & 14,715 & 9,435 & 5,280 & 209,350 & 28.5 & 0.19\% \\
        \midrule
        \multirow{3}{*}{Test}
          & code  &  6,307 & 4,043 & 2,264 &  94,160 & 29.9 & 0.47\% \\
          & lab   &  6,307 & 4,043 & 2,264 &   4,110 &  1.3 & 0.02\% \\
          & image &  6,307 & 4,043 & 2,264 &  88,360 & 28.0 & 0.44\% \\
        \bottomrule
    \end{tabular}
\end{table}
% \begin{table}[h!]
%     \centering
%     \caption{Edge homophily. For each similarity edge we measure whether the two connected visits share a clinical attribute, relative to a random-edge null (lift = observed / null).}
%     \label{tab:edge_homophily}
%     \begin{tabular}{lccc}
%         \toprule
%         \textbf{Attribute} & \textbf{Observed} & \textbf{Null} & \textbf{Lift} \\ \midrule
%         Same primary ICD chapter & 0.370 & 0.123 & 3.00$\times$ \\
%         Same one-year mortality outcome & 0.700 & 0.643 & 1.09$\times$ \\
%         Same 30-day readmission outcome & 0.697 & 0.685 & 1.02$\times$ \\ \bottomrule
%     \end{tabular}
% \end{table}

\section*{S6 Supplementary Results: Graph-Free Ablation}

To establish the necessity of cross-patient graph construction over standard multimodal fusion, we evaluate a graph-free control variant in which the GNN modules are removed entirely. In this variant, the initial modality embeddings are mapped directly into the shared representation space, while the fusion, alignment, and context-injection components remain identical to the full model and are trained under the same procedure. This design isolates cross-patient graph construction as the only difference between the two models.

\ref{tab:graph_free} reports the results. Removing the graph degrades AUC on both tasks (mortality: $0.702 \rightarrow 0.685$; readmission: $0.548 \rightarrow 0.527$), indicating that the performance of \proposed cannot be attributed to multimodal fusion and LLM injection alone, and that propagating information across the patient-visit graph contributes materially to predictive performance.

\begin{table}[h!]
    \centering
    \caption{Graph-free ablation. AUC (mean $\pm$ standard deviation) for the full model and a graph-free variant in which GNN modules are replaced by a direct mapping of initial embeddings into the shared space.}
    \label{tab:graph_free}
    \begin{tabular}{lcc}
        \toprule
        \textbf{Task} & \textbf{M$^2$G-LLM (with graph)} & \textbf{Graph-free} \\ \midrule
        One-year mortality & \textbf{0.702} $\pm$ 0.018 & 0.685 $\pm$ 0.031 \\
        30-day readmission & \textbf{0.548} $\pm$ 0.023 & 0.527 $\pm$ 0.014 \\ \bottomrule
    \end{tabular}
\end{table}

\section*{S7 Supplementary Analysis: Graph Homophily}
\label{sec:S7}

To assess whether the constructed similarity graphs capture clinically
meaningful structure, and to verify that they provide no shortcut to the
prediction target, we measure edge homophily with respect to three attributes:
primary ICD chapter, one-year mortality label, and 30-day readmission label.
Edge homophily is defined as the fraction of similarity edges joining two visits
that share the attribute in question. We compare this against a null model in
which edges are rewired at random among cross-patient visit pairs with matched
degree, and report lift as the ratio of observed to null homophily. A lift of
1.0 indicates that edges are no more likely to join visits sharing the attribute
than chance. Temporal edges are excluded, as they join visits of the same
patient by construction. The analysis is performed independently on the training
and test graphs.

\begin{table}[htbp]
  \centering
  \caption{Edge homophily of similarity graphs, reported as lift over a
    degree-matched cross-patient random-edge null. Lift is the ratio of observed
    to null homophily; 1.0 indicates chance. Temporal edges are excluded.}
  \label{tab:S13}
  \begin{tabular}{llccc}
    \toprule
    Split & Modality & ICD-chapter lift & Mortality lift & Readmission lift \\
    \midrule
    Train & code  & $2.82\times$ & 0.99 & 1.00 \\
    Train & lab   & $1.63\times$ & 0.99 & 1.01 \\
    Train & image & $1.17\times$ & 1.02 & 0.99 \\
    Test  & code  & $2.68\times$ & 1.00 & 1.01 \\
    Test  & lab   & $1.78\times$ & 0.89 & 0.95 \\
    Test  & image & $1.13\times$ & 1.02 & 0.98 \\
    \bottomrule
  \end{tabular}
\end{table}

Two findings follow. First, the graphs are diagnosis-homophilous, and this is
driven principally by the codified modality: code-graph edges connect visits
sharing a primary ICD chapter at $2.8\times$ chance, replicating closely on the
held-out split ($2.7\times$). The laboratory graph is moderately homophilous
($1.6$--$1.8\times$), while the imaging graph is only marginally above chance
($1.1\times$). The weak structure of the imaging graph is consistent with the
concentration of its underlying similarity distribution: with a median
inter-patient cosine of 0.988 between ResNet50-ImageNet embeddings, top-$k$
selection operates on differences that are small relative to noise, and the
resulting edges carry limited semantic content. We therefore do not claim that
all three modality graphs encode clinically meaningful similarity to the same
degree, and identify a domain-pretrained imaging encoder as a natural
improvement.

Second, none of the graphs is outcome-homophilous. All mortality and readmission
lifts fall between 0.89 and 1.02, i.e.\ indistinguishable from chance. This is
the expected consequence of a construction that never observes labels, and it
provides direct empirical evidence that the graph offers no shortcut to the
prediction target, complementing the structural argument given in Methods that
no edge crosses the train--test split. Predictive benefit instead arises at the
diagnostic level: message passing aggregates over patients with related clinical
presentations, consistent with the graph-free control (Supplementary
Table~12), where removing propagation reduces one-year mortality AUC from 0.702
to 0.685 with all modalities present.

\section*{S8 Supplementary Results: Sensitivity to Injection Strength}
\label{sec:S8}

The residual-stream injection is governed by two scalar coefficients,
$\lambda_{\mathrm{mha}}$ and $\lambda_{\mathrm{mlp}}$. To characterize
the model's sensitivity to injection strength, we sweep
$\lambda_{\mathrm{mha}} \in \{0, 0.10, 0.25, 0.50, 1.0\}$ with
$\lambda_{\mathrm{mlp}} = 1.0$ held fixed, retraining the framework at each
setting and reporting AUC on both tasks. Results are averaged over three seeds.

\begin{table}[htbp]
  \centering
  \caption{Sensitivity to injection strength. AUC (mean $\pm$ standard deviation
    over three seeds) as a function of $\lambda_{\mathrm{mha}}$, with
    $\lambda_{\mathrm{mlp}} = 1.0$ fixed.}
  \label{tab:S14}
  \begin{tabular}{lcc}
    \toprule
    $\lambda_{\mathrm{mha}}$ & Mortality AUC & Readmission AUC \\
    \midrule
    0    & $0.589 \pm 0.018$ & $0.523 \pm 0.040$ \\
    0.10 & $0.702 \pm 0.018$ & $0.548 \pm 0.023$ \\
    0.25 & $0.703 \pm 0.043$ & $0.542 \pm 0.025$ \\
    0.50 & $0.718 \pm 0.021$ & $0.537 \pm 0.035$ \\
    1.0  & $0.693 \pm 0.030$ & $0.525 \pm 0.037$

 \\
    \bottomrule
  \end{tabular}
\end{table}

Injection strength is a consequential hyperparameter. At
$\lambda_{\mathrm{mha}} = 0$ the model performs near chance on both tasks
(mortality AUC 0.589), close to the text-only Llama-3-8B baseline (0.558,
Table~1), confirming that the predictive signal is carried by the injected
multimodal context vector rather than by the clinical text prompt alone.
Among the non-zero settings, $\lambda_{\mathrm{mha}} = 0.50$ attains the
highest mortality AUC (0.718), but readmission AUC declines monotonically
beyond $\lambda_{\mathrm{mha}} = 0.10$; we therefore retain
$\lambda_{\mathrm{mha}} = 0.10$ for the main framework as the setting that
maximizes readmission AUC while remaining within one standard deviation of
the best mortality result, providing the best balance across both tasks.
\section*{S9 Supplementary Results: Robustness to Inference-Time Modality Missingness}
\label{sec:S9}

The modality ablations in Table~3 and Supplementary Table~8 evaluate performance
when a modality is withheld from the model entirely during training. This is
distinct from the deployment-relevant case in which a modality is unavailable
for a subset of patients at prediction time. We therefore evaluate
inference-time missingness directly.

Using trained models without modification, each structured modality --- diagnosis
codes, laboratory values, and imaging --- is dropped independently and per patient
with probability $p \in \{0, 0.2, 0.5, 0.8\}$. Dropout masks are held identical
across compared models. Missingness is applied at the GNN input, so that message
passing may reconstruct an absent modality from clinically similar neighbours
(Methods, Multimodal Alignment), whereas a model without graph propagation
cannot. Results are averaged over three seeds; $p = 0$ anchors each curve. We
compare against mmFormer, the strongest non-LLM baseline on this task, and HeLM,
the pseudo-token LLM baseline whose input-stream injection strategy
M\textsuperscript{2}G-LLM is designed to contrast with.

\begin{table}[htbp]
  \centering
  \caption{One-year mortality AUC under inference-time modality dropout. Each
    structured modality is dropped independently per patient with probability
    $p$, using trained models without modification. Mean $\pm$ standard
    deviation over three seeds.}
  \label{tab:S15}
  \begin{tabular}{lcccc}
    \toprule
    Model & $p = 0$ & $p = 0.2$ & $p = 0.5$ & $p = 0.8$ \\
    \midrule
    M\textsuperscript{2}G-LLM & $0.702 \pm 0.018$ & $0.698 \pm 0.014$ & $0.682 \pm 0.014$ & $0.669 \pm 0.013$ \\
    mmFormer                  & $0.693 \pm 0.026$ & $0.627 \pm 0.025$ & $0.603 \pm 0.025$ & $0.587 \pm 0.025$ \\
    HeLM                      & $0.554 \pm 0.023$ & $0.550 \pm 0.009$ & $0.548 \pm 0.019$ & $0.524 \pm 0.017$ \\
    \bottomrule
  \end{tabular}
\end{table}

M\textsuperscript{2}G-LLM degrades by 3.3 AUC points from $p = 0$ to $p = 0.8$,
compared with 10.6 points for mmFormer. HeLM shows a smaller absolute decrement,
but from a substantially lower baseline and with only one droppable structured
modality, and remains below M\textsuperscript{2}G-LLM at every level of
missingness.

\section*{S10 Supplementary Results: GNN Depth and Architecture}
\label{sec:S10}

We evaluate the effect of propagation depth and of the graph convolution
operator on one-year mortality AUC, comparing GCN against GraphSAGE and GAT at
depths of one, two, and three layers. All other components of the framework are
held fixed. Results are averaged over three seeds.

\begin{table}[htbp]
  \centering
  \caption{One-year mortality AUC by graph convolution operator and propagation
    depth. Mean $\pm$ standard deviation over three seeds.}
  \label{tab:S16}
  \begin{tabular}{lccc}
    \toprule
    Architecture & Depth 1 & Depth 2 & Depth 3 \\
    \midrule
    GCN       & $0.664 \pm 0.014$ & $0.686 \pm 0.006$ & $0.702 \pm 0.018$ \\
    GraphSAGE & $0.661 \pm 0.029$ & $0.673 \pm 0.003$ & $0.701 \pm 0.002$ \\
    GAT       & $0.675 \pm 0.007$ & $0.687 \pm 0.003$ & $0.705 \pm 0.003$ \\
    \bottomrule
  \end{tabular}
\end{table}

Performance improves with depth for all three operators, consistent with
progressively wider neighbourhood aggregation. Differences between operators at
a given depth are small relative to their dispersion, with GAT marginally ahead
at each depth; we retain GCN for the main framework on the grounds of parsimony
and lower computational cost, as the attention parameters of GAT and the
sampling machinery of GraphSAGE do not yield a commensurate improvement on these
graphs.

\section*{S11 Supplementary Analysis: Computational Cost}
\label{sec:S11}

All experiments were conducted on a single NVIDIA B200 GPU. Training requires
492\,s per epoch over approximately five epochs. Inference costs 154\,ms per
patient. The Llama-3-8B backbone occupies 16.3\,GB in bfloat16, with peak memory
reaching 36.1\,GB during a 4096-token forward pass with injection. The framework
trains 53.11\,M parameters, 0.66\% of the 8.03\,B total, as the language backbone
remains frozen throughout.

Similarity-edge construction requires $O(N^2 d)$ operations for $N$ visits and
$d$-dimensional embeddings. The binding constraint in practice is memory rather
than compute: the dense $N \times N$ similarity matrix requires approximately
866\,MB in float32 at the present cohort size and grows quadratically. This cost
is incurred once offline; graph topology is fixed thereafter, and no similarity
computation occurs during training or inference.